\documentclass[draft=false,sigconf]{acmart}
\pdfoutput=1
\usepackage{natbib}
\usepackage{xcolor}
\usepackage{booktabs} % For formal tables
\usepackage{enumitem}

\setcopyright{none}

\begin{document}
\title[Efficient Hybrid Network Architectures for Extremely Quantized Neural Networks]{Efficient Hybrid Network Architectures for Extremely Quantized Neural Networks Enabling Intelligence at the Edge}
%\titlenote{Produces the permission block, and
%  copyright information}
%\subtitle{Extended Abstract}
%\subtitlenote{The full version of the author's guide is available as
%  \texttt{acmart.pdf} document}

\author{Indranil Chakraborty, Deboleena Roy, Aayush Ankit and Kaushik Roy}
\affiliation{%
  \institution{Purdue University}
  \streetaddress{465 Northwestern Ave}
  \city{West Lafayette}
  \state{Indiana}
  \postcode{47907}
}
\email{ichakra@purdue.edu, roy77@purdue.edu, aankit@purdue.edu, kaushik@purdue.edu}

%\author{Deboleena Roy}
%\affiliation{%
%  \institution{Purdue University}
%  \streetaddress{465 Northwestern Ave}
%  \city{West Lafayette}
%  \state{Indiana}
%  \postcode{47907}
%}
%\email{roy77@purdue.edu}

%\author{Kaushik Roy}
%\affiliation{%
% \institution{Purdue University}
%  \streetaddress{465 Northwestern Ave}
%  \city{West Lafayette}
%  \state{Indiana}
%  \postcode{47907}}
%\email{kaushik@purdue.edu}


\begin{abstract}
The recent advent of `Internet of Things' (IOT) has increased the demand for enabling AI-based edge computing. This has necessitated the search for efficient implementations of neural networks in terms of both computations and storage. Although extreme quantization has proven to be a powerful tool to achieve significant compression over full-precision networks, it can result in significant degradation in performance. In this work, we propose extremely quantized hybrid network architectures with both binary and  full-precision sections to emulate the classification performance of full-precision networks while ensuring significant energy efficiency and memory compression. We explore several hybrid network architectures and analyze the performance of the networks in terms of accuracy, energy efficiency and memory compression. We perform our analysis on ResNet and VGG network architectures. Among the proposed network architectures, we show that the hybrid networks with full-precision residual connections emerge as the optimum by attaining accuracies close to full-precision networks while achieving excellent memory compression, up to 21.8x in case of VGG-19. This work demonstrates an effective way of hybridizing networks which achieve performance close to full-precision networks while attaining significant compression, furthering the feasibility of using such networks for energy-efficient neural computing in IOT-based edge devices.
\end{abstract}


%\ccsdesc[500]{Computer systems organization~Embedded systems}


\maketitle

\vspace{-4mm}
\section{Introduction}

Deep Learning has emerged as the most instrumental cog in revolutionizing the current drive of ubiquitous Artificial Intelligence (AI), especially in domains of visual recognition and natural language processing. Over the past decade, extensive research in deep learning has enabled machines to go beyond image classification \cite{krizhevsky2012imagenet,szegedy2015going,he2016deep,girshick2015fast} and natural language processing \cite{mikolov2013efficient} to the extent of outperforming humans in games such as Atari \cite{mnih2015human} and Go \cite{silver2017mastering}. Despite the unprecedented success of deep neural networks, standard network architectures prove to be intensive both in terms of memory and computational resources and require expensive GPU-based platforms for execution. However, with the advent of the modern age of `Internet of Things' and a proliferating need for enabling AI in low-power edge devices, designing energy and memory-efficient neural networks is quintessential. This has driven researchers to look at ways to reduce model complexity, while trying to meet the algorithmic requirements, like accuracy and reliability. 

One way to reduce model size is to modify the network architecture itself, such that it has fewer parameters. SqueezeNet \cite{iandola2016squeezenet} employs a series of 1$\times$1 convolutions to compress and expand feature maps as they pass through the neural network. Another method of compression is pruning which aims to reduce redundancies in over-parameterized networks. Hence, researchers have investigated several network pruning techniques, both during training \cite{alvarez2017compression, weigend1991generalization} and inferencing \cite{han2015learning, ullrich2017soft}. 

A different technique of model compression is reduced bit precision to represent weights and activations. Quantizing networks achieves energy efficiency and memory compression compared to full-precision networks. 
%Binary networks, especially, can further leverage circuit techniques for performing low-precision computations in memory \cite{liu2018parallelizing}, thus attempting to alleviate the von Neumann memory bottleneck. 
Several training algorithms have been proposed to train such binary and ternary neural networks \cite{hubara2017quantized,mellempudi2017ternary}. Although these algorithms attain close to performance of a full-precision network for smaller datasets such as MNIST and CIFAR-10, scaling them to ImageNet is a challenge. As a solution, XNOR-Nets (binary weights and activations) and BWNs (binary weights and full-precision activations) \cite{rastegari2016xnor} were proposed. They offer a different scheme of binarization that uses a scaling factor per weight filter bank and were able to scale to ImageNet, albeit, with a degradation in accuracy compared to a full-precision network. Researchers have also looked at a hybrid network structure combining BWNs and XNOR-Nets \cite{prabhu2018hybrid}, where activations are full precision for certain layers and binary for others. However, despite such hybridization techniques, the gulf between full precision networks and quantized networks still remain considerably high, especially for deep networks and larger datasets. In light of these shortcomings of quantization algorithms, we propose hybrid network architectures combining binary and full-precision sections to attain performance close to full-precision networks while achieving significant energy efficiency and memory compression. We explore several hybrid networks which involve adding full-precision residual connections, and breaking the network into binary and full-precision sections, both layer-wise and within a layer. We evaluate the performance of the proposed networks on datasets such as CIFAR-100 and ImageNet and explore the trade-offs between classification accuracy, energy efficiency and memory compression. We compare the different kinds of hybrid network architectures to identify the optimum network which recover the performance degradation in extremely quantized networks while achieving excellent compression. Our approach provides an effective way of designing hybrid networks which attain the classification performance close to full-precision networks while achieving significant compression, thereby increasing the feasibility of ubiquitous use of low-precision networks in low-power edge devices.

\section{Design Methodology for Hybrid Networks}
We address the problem of performance degradation due to extreme quantization by proposing hybrid network architectures constituted by binary networks with full-precision elements in different forms. Introducing full-precision elements could be in form of adding full-precision residual layers or might involve splitting the neural networks into binary and full-precision sections. A full-precision layer here means both full-precision weights and activations unless mentioned otherwise. We use the binarization scheme developed in \cite{rastegari2016xnor} as they have been demonstrated to scale to large datasets. 
The binary convolution operation between inputs $X$ and weights $W$ is approximated as:
\begin{equation}
	X*W \approx (sign(X)*sign(W))\odot\alpha  
\end{equation}
Here, $\alpha$ is the L1-norm of $W$. These binary convolutions are similar to XNOR operations and hence these networks are called XNOR-Nets. We define XNOR-Net as our baseline binary network architecture. As in \cite{rastegari2016xnor}, we have kept the first and final layers of hybrid networks full-precision. We apply our hybridization techniques on the binary layers of XNOR-Net. The types of hybrid network architectures explored are described below:
\vspace{-1mm}
\begin{enumerate}
	\item \textbf{\textit{Hybrid Networks with full-precision residual connections:}} This kind of hybrid networks are comprised of binary networks along with a few full-precision residual connections. Residual connections are usually unity connections which run parallel to the convolutional layers and are added to the convolution output every few layers. Some of the residual connections might have weight filters to downsample the input maps when the filter size changes. As these weight layers are computationally less intensive, making them full-precision promises improvement of classification accuracies while still achieving a large compression compared to full-precision networks. For networks that do not have residual connections, we add a few full-precision connections to form a hybrid network. 
	\item \textbf{\textit{Sectioned Hybrid Networks}}: We explore an alternative technique to hybridize networks with network sectioning. Sectioning the network involves splitting the network into binary and full-precision sections. Sectioning can be of two forms: 
	\begin{enumerate}
		\item  \textbf{\textit{Inter-layer sectioning}}: This type of sectioning involves splitting the neural network with $N$ layers into $k$ binary and $N-k$ full-precision layers. 
		\item  \textbf{\textit{Intra-layer sectioning}}: This type of sectioning involves splitting each layer of the neural network into binary and full-precision sections with a fraction $p$, which means $p/100$ \% of the weight filters of each layer are full-precision. 
	\end{enumerate}
	
\end{enumerate}
In our analysis, we consider ResNet and VGG network architectures. For ResNets, we also analyze all the proposed hybrid ResNet-based networks for different network widths. However, as VGG-Nets are inherently over-parameterized, we only explore the hybrid VGG networks for one particular network width. 

\subsection{Hybrid ResNets with full-precision residual connections}
We consider the network architecture ResNet-N (a ResNet with N layers) which has $N-1$ convolutional and 1 fully-connected weight layers. We propose a hybrid ResNet architecture constituted by a XNOR ResNet along with full-precision residual connections. The residual connections in ResNets are usually identity connections except in cases where the connection needs to be downsampled with $1\times1$ convolution kernels. The network proposed using full-precision residual connections is described below:

\textbf{\textit{Hybrid-Res A: }} This configuration is comprised of a binary ResNet along with full-precision residual convolutional layers. Fig. \ref{fig:resnet} (a) shows this architecture where the full-precision downsampling residual layers are shown in color. 
\vspace{-4mm}
\subsection{Sectioned Hybrid ResNets}	

\begin{figure}[t]
	\centering
	\includegraphics[width=3.6in, keepaspectratio]{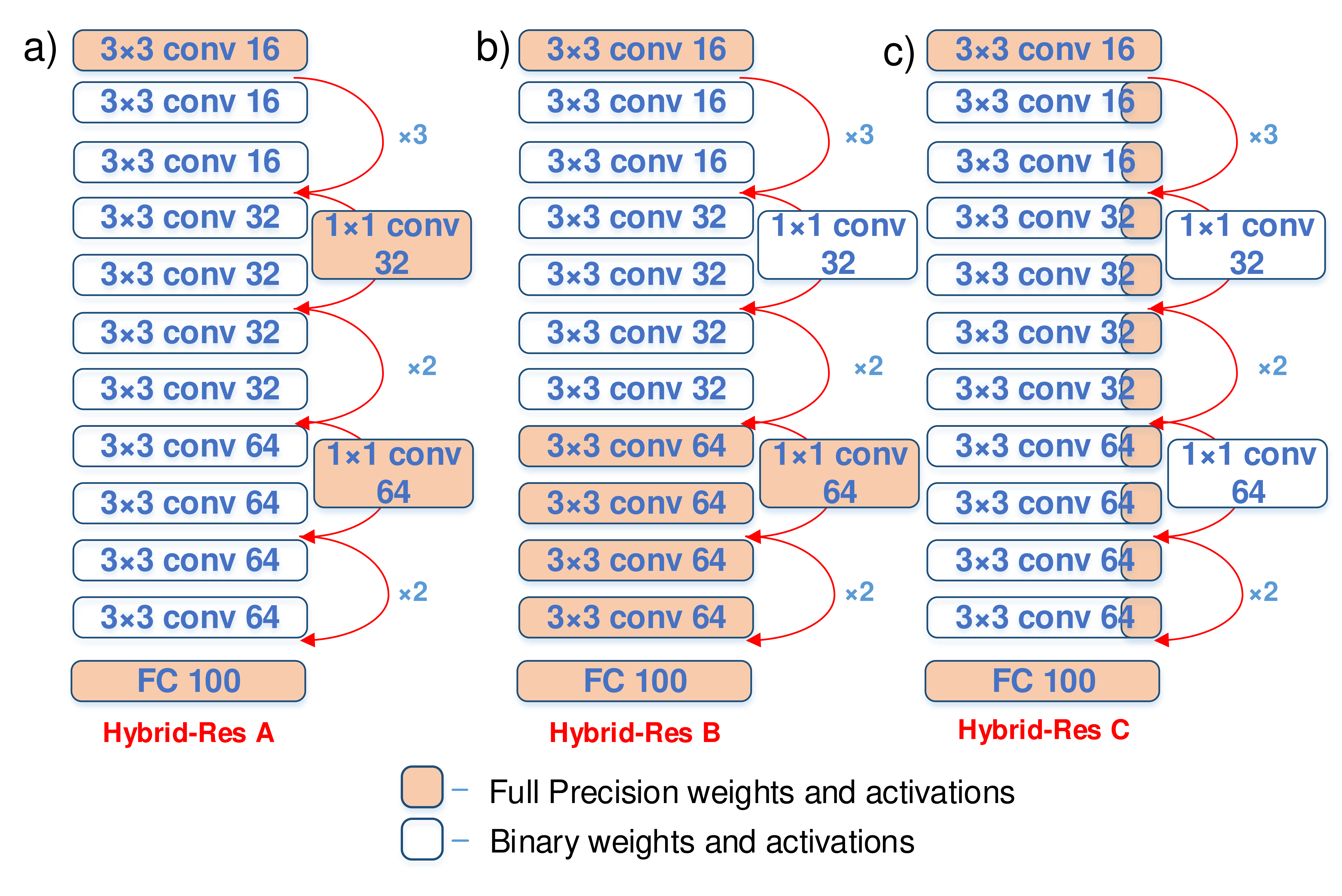}
	\vspace{-6mm}
	\caption{Three configurations of hybrid network architectures based on ResNet-20 (as an example), (a) Hybrid-Res A, (b) Hybrid-Res B and (c) Hybrid-C. The proposed architecture Hybrid-Res A is a ResNet where the residual convolutional layers are full-precision (colored). Hybrid-Res B and Hybrid-Res C are sectioned hybrid networks where we split the ResNet into binary(uncolored) and full-precision (colored) sections both inter-layer and intra-layer respectively.}
	\label{fig:resnet}
	\vspace{-6mm}
\end{figure}

We also explore an alternative technique to hybridize by breaking the neural networks into binary and full-precision sections. This can be done in two ways, inter-layer and intra-layer. The networks are described below:
\begin{enumerate}
	\item \textbf{\textit{Hybrid-Res B: }}   In this configuration, we perform inter-layer sectioning where we section the ResNet into $k$ binary layers and $N-k$ full precision layers. For example, in ResNet-20, this sectioning can lead to 6 full-precision layers and 12 binary layers, as shown in Fig. \ref{fig:resnet} (b).
	\item \textbf{\textit{Hybrid-Res C: }}  In This configuration, we split each layer of the ResNet into binary and full-precision sections with a fraction $p$ as defined earlier (shown in Fig. \ref{fig:resnet} (c)).
\end{enumerate}

\subsection{Hybrid VGG Networks with full-precision residual connections}
\begin{figure}[t]
	\centering
	\includegraphics[width=3.2in, keepaspectratio]{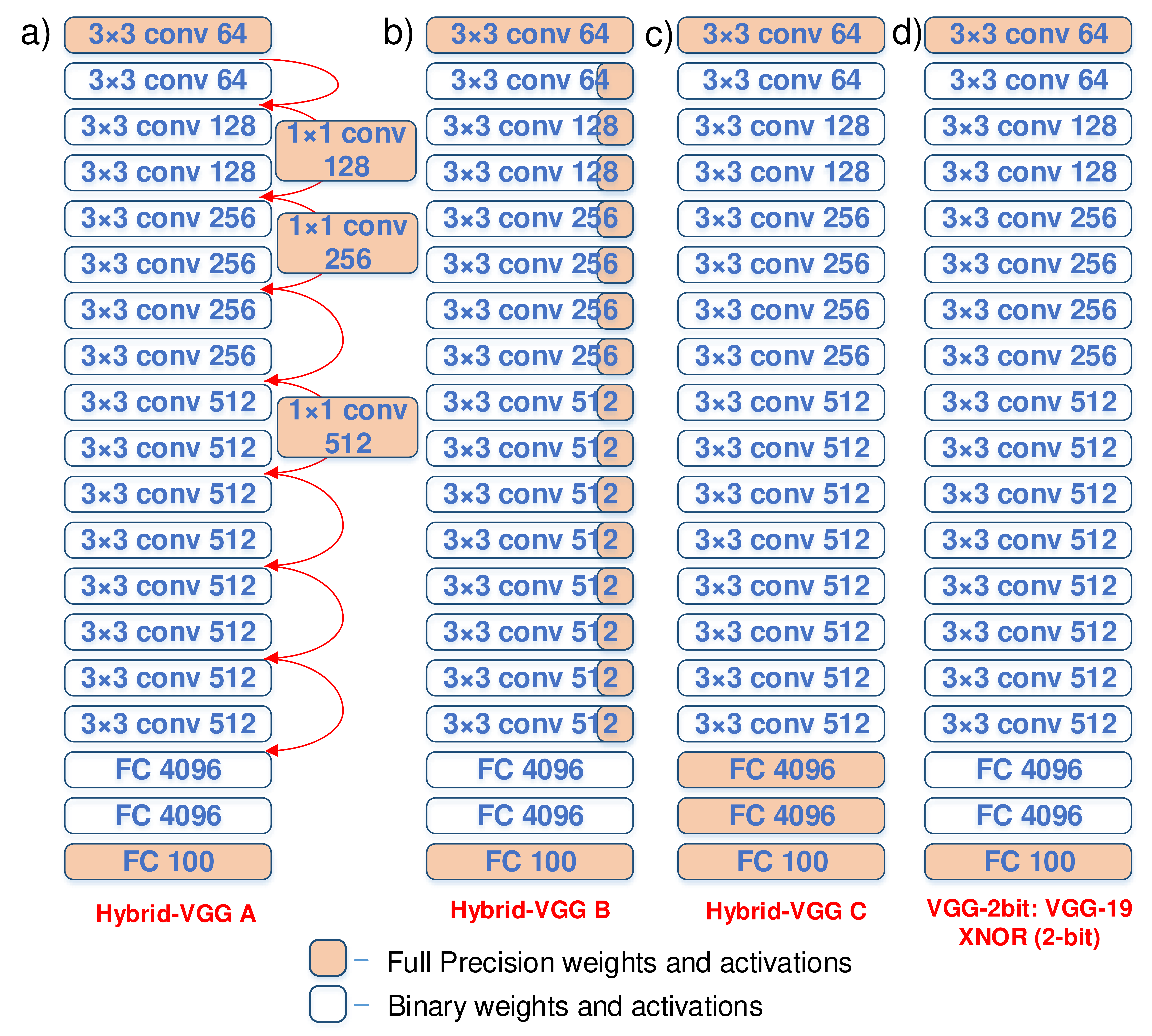}
	\vspace{-4mm}
	\caption{(a) Hybrid-VGG A is based on VGG-N architecture (as an example) with a few full-precision convolutional residual connections. Its variant Hybrid-VGG A' has the second FC 4096 layer full-precision. (b) Hybrid-VGG B has intra-layer sectioning, i.e., partially full-precision weights in each layer. (c) Hybrid-VGG C has inter-layer sectioning, i.e, split the network into binary and full-precision sections. (d) VGG-2bit is basic XNOR-Net structure where weights are 2-bit. FC is fully connected. }
	\label{fig:vgg-res}
	\vspace{-3mm}
\end{figure}

We consider the network architecture VGG-N which has $N-3$ convolutional kernel layers and 3 fully connected layers. We propose a hybrid VGG network design by adding full-precision residual connections to the standard VGG network. We extend that concept of using full-precision residual connections to VGG networks. The network is described below and depicted in Fig. \ref{fig:vgg-res} (a):

\textbf{\textit{Hybrid-VGG A:}} In this network, we add residual connections to the standard VGG-N architecture every two convolutional layers. Each time the number of output maps changes, we include a full-precision downsampling $1\times1$ convolutional layer in the residual path, as shown in color in Fig. \ref{fig:vgg-res} (a). For ImageNet simulations, we made the second fully connected layer of this configuration full-precision. Let us call this network Hybrid-VGG A'. 

\subsection{Sectioned Hybrid VGG Networks}
We explore sectioned hybridization techniques involving breaking the VGG network into binary and full-precision sections. We consider inter-layer and intra-layer sectioning. The descriptions of the networks are listed below and depicted in Fig. \ref{fig:vgg-res} (b) and (c):   \par 
\begin{enumerate}	
	\item \textbf{\textit{Hybrid-VGG B:}} In this network, we consider intra-layer sectioning where we make a fraction $p$ of each convolutional layer full-precision. The full-precision sections are shown in color in Fig. \ref{fig:vgg-res} (b). 
	\item \textbf{\textit{Hybrid-VGG C:}} This network considers inter-layer sectioning involving $k$ full-precision layers and $N-k$ binary layers. For example, in VGG-19 architecture (Fig. \ref{fig:vgg-res} (c)), we can split the binary section of the XNOR-Net into 15 binary convolutional layers and 2 full-precision linear layers.
\end{enumerate}
For VGG networks, we further compare the proposed hybrid networks with networks with increased widths and increased bit-precision. These networks are based on the basic VGG XNOR-Net, shown in Fig. \ref{fig:vgg-res} (d). The networks are described below:
\begin{enumerate}
	\item \textbf{\textit{VGG-2bit:}} We increase the bit-precision of weights of the quantized layers to 2-bit from 1-bit in case of XNOR-Nets that we have explored thus far. The training algorithm for any k-bit quantized network can be readily derived from XNOR-Net \cite{rastegari2016xnor} where the quantized levels are: 
	
	$q_k(x) = 2(\frac{\lfloor{(2^k-1)(x+1)/2}\rfloor}{2^k-1}-\frac{1}{2})$
	instead of the $sign$ function, as in case of binary networks. 
	\item \textbf{\textit{VGG-Inflate:}} We increase the network width 2 times, i.e, the number of filter banks in each convolutional layer is doubled. 
\end{enumerate}

\section{Experiments, Results and Discussion}

\subsection{Energy and Memory Calculations}
We perform a storage and computation analysis to calculate the energy efficiency and memory compression of the proposed networks. For any two networks A and B, the energy efficiency of Network A with respect to Network B can be defined as: Energy Efficiency (E.E) = (Energy consumed by Network A)/(Energy consumed by Network B). Similarly, Memory compression of Network A with respect to Network B can be defined as: Memory Compression (M.C) = (Storage required by Network A)/(Storage required by Network B). E.E (FP) (M.C (FP)) is the energy efficiency (memory compression) of a network with respect to the full-precision network whereas E.E (XNOR) (M.C (XNOR)) is the energy efficiency (memory compression) of the network with respect to the XNOR network. 
\subsubsection{\textbf{Energy Efficiency}}
We considered the energy consumed by the computations (multiply-and-accumulate or MAC operations) and memory accesses in our calculations for energy efficiency. We do not take into account energy consumed due to data flow and instruction flow in the architecture. For a convolutional layer, there are $I$ input channels and $O$ output channels. Let the size of the input be $N\times N$, size of the kernel be $k \times k$ and size of the output be $M\times M$. Thus, in Table \ref{ops} we present the number of memory-accesses and computations for standard full-precision networks:
\vspace{-3mm}
\begin{table}[h!]
	\caption{Operations in neural networks}
	\vspace{-3mm}
	\label{ops}
	\begin{tabular}{|l|l|}
		\hline
		Operation          & Number of Operations               \\ \hline\hline
		Input Read         & $N^2\times I$                      \\ \hline
		Weight Read        & $k^2\times I\times O$              \\ \hline
		Computations (MAC) & $M^2 \times I \times k^2 \times O$ \\ \hline
		Memory Write       & $M^2\times O$                      \\ \hline
	\end{tabular}
\end{table}

The number of binary operations in the binary layers of the hybrid networks is same as the number of full-precision operations in the corresponding layers of the full-precision networks. Since we use the XNOR-Net training algorithm for training the binary weights in our work, we consider additional full-precision memory accesses and computations for parameter $\alpha$, where $\alpha$ is the scaling factor for each filter bank in a convolutional layer. Number of accesses for $\alpha$ is equal to the number of output maps, $O$. Number of full-precision computations are $M^2\times O$.  

We calculated the energy consumption from projections for 10 nm CMOS technology \cite{keckler2011gpus}. Considering 32-bit representation as full-precision, the energy consumption for both binary and 32-bit memory accesses and computations are shown in Table. \ref{energy}.
\vspace{-3mm}
\begin{table}[h!]
	\caption{Energy Consumption chart}
	\vspace{-3mm}
	\label{energy}
	\begin{tabular}{|p{1.5cm}|l|p{1.7cm}|l|}
		\hline
		Operation                    & Energy (pJ) & Operation                    & Energy (pJ)\\ \hline\hline
		%	Binary Memory Access         & 2.5         \\ \hline
		%	Binary MAC                   & 0.1         \\ \hline
		32-bit Memory Access & 80    & Binary Memory Access & 2.5      \\ \hline
		32-bit MAC           & 3.25  & Binary MAC & 0.1       \\ \hline
	\end{tabular}
\end{table}
\vspace{-3mm}
The energy numbers for binary memory accesses and MAC operations are scaled down 32 times from the corresponding full-precision values. Let the number of full-precision (binary) memory accesses in any layer be $ME_{FP}$ ($ME_{Bin}$) and number of full-precision (binary) computations in any layer be
$C_{FP}$ ($C_{Bin}$). Then, energy consumed by any layer is given by $E = ME_{FP}\times80+ME_{Bin}\times2.5+C_{FP}\times3.25+C_{Bin}\times0.1$. For a binary layer, $ME_{FP} = O, C_{FP} = M^2 O$ and $ME_{Bin} = (N^2I+k^2I+M^2O), C_{Bin} = M^2Ik^2O$. For a full-precision layer, $ME_{FP} = (N^2I+k^2I+M^2O), C_{FP} = M^2Ik^2O$.  Note, this calculation is a rather conservative estimate which does not take into account other hardware architectural aspects such as input-sharing or weight-sharing. However, our approach concerns with modifications of network architecture and we compare the ratios of energy consumption. These aspects of the hardware architecture affect all the networks equally and hence can be taken out of consideration.

%The basic principle for energy calculation of any layer is \textit{E = }. For example, energy consumed by a full-precision layer is $E_{FP} = ((N^2\times I+k^2\times I+M^2\times O)\times80 + M^2 \times I \times k^2 \times O \times 3.25) $ pJ whereas energy consumed by a binary layer would be $E_{Bin} = ((N^2\times I+k^2\times I+M^2\times O)\times2.5 + O\times 80 + M^2 \times I \times k^2 \times O \times 0.1 + M^2\times O\times 3.25) $ pJ. 
%\begin{table}[h!]
%	\caption{Number of full-precision operations in XNOR-Nets}
%	\label{binop}
%	\begin{tabular}{|l|l|}
%		\hline
%		Operations                 & Number of Operations \\ \hline\hline
%		$\alpha$ read              & O                    \\ \hline
%		Full-Precision computation & $M^2 \times O$       \\ \hline
%	\end{tabular}
%\end{table}
\subsubsection{\textbf{Memory Compression}}
The memory required for any network is given by product of the total number of weights in the network multiplied by the precision of the weights. The number of weights in each layer is given by $N_w = I\times O\times k^2$. Thus, the memory required by a full-precision layer is $N_w\times 32$ bits and that of a binary layer is $N_w$ bits.

Note that the assumption for the energy and storage calculations for binary layers hold for custom hardware capable of handling fixed-point binary representations of data, thus leveraging the benefits offered by quantized networks. 
\subsection{Image Classification Framework}
We evaluated the performance of all the networks described in this section in PyTorch. We explore hybridization of 2 network architectures, namely, ResNet-20 and VGG-19 where the training algorithm for the binarized layers has been adopted from `XNOR-Net' \cite{rastegari2016xnor}. We perform image classification on the dataset CIFAR-100 \cite{krizhevsky2009learning}. The CIFAR-100 dataset has 50000 training images and 10000 testing images of size $32\times32$ for 100 classes. We report classification performance using top-1 test accuracies for the datasets. 
\subsection{Drawbacks of XNOR-Nets}
Firstly, we evaluate the performance of our baseline binary networks or XNOR-Nets for VGG-19 and ResNet-20. This analysis will help us understand the drawbacks of using purely binary networks without any hybridization. Table. \ref{XNORBL} lists the accuracy, energy effiency and memory compression of XNOR-Nets:
	\vspace{-4mm}
\begin{table}[h!]
	\caption{Comparison of XNOR VGG-19 and ResNet-20 on CIFAR-100}
	\vspace{-4mm}
	\label{XNORBL}
	\resizebox{0.5\textwidth}{!}{%
		\begin{tabular}{|l|p{2cm}|p{1.5cm}|p{1.5cm}|p{1.5cm}|}
			\hline
			Network   & Full-Precision Accuracy (\%) & XNOR-Net Accuracy (\%) & E.E (FP) & M.C (FP) \\ \hline
			
			VGG-19    & 67.21                   & 37.47    & 24.13            & 24.08             \\ \hline
			Resnet-20 & 65.81                   & 50.2     & 18.67             & 17.26             \\ \hline
		\end{tabular}%
	}
\vspace{-4mm}
\end{table}

We observe that VGG-19 and ResNet-20 have similar full-precision accuracies, however, XNOR-Net VGG-19 suffers a significantly higher degradation in accuracy compared to XNOR-Net ResNet-20 on CIFAR-100. Inflating the networks, i.e., making the networks wider is a way to improve upon the degradation in accuracy suffered by XNOR-Nets. As shown in Table. \ref{inflate}, inflating the networks improves the accuracy of XNOR ResNet-20 close to full-precision accuracy at the cost of memory compression, however, in case of XNOR VGG-19, the improvement is not significant. To address this degradation in accuracy, we propose hybrid network architectures where we use a few full-precision layers in extremely quantized networks to improve the performance of XNOR-Nets while still achieving significant energy-efficiency and memory compression with respect to full-precision networks.
\vspace{-3mm}
\begin{table}[h!]
	\caption{Impact of width on XNOR-Net on CIFAR-100}
	\vspace{-0.3cm}
	\text{Full-precision accuracy: 65.81\%}
	\label{inflate}
	\resizebox{0.5\textwidth}{!}{%
		\begin{tabular}{|p{2cm}|l|p{1cm}|p{1.5cm}|p{1.1cm}|p{1.6cm}|}
			\hline
			\multicolumn{6}{|c|}{\textbf{ResNet-20}} \\ \hline
			Network Width          & Accuracy  & E.E (FP) & E.E (XNOR) & M.C (FP) & M.C (XNOR) \\ \hline
			1x (ResNet XNOR-Net) & 50.2       & 18.67   &  1        & 17.26     &   1    \\ \hline
			1.5x                   & 57.86        & 10.09 &  0.54           & 9.1    & 0.52          \\ \hline
			2x                     & 63.09      & 6.45    &  0.35        & 5.65     &  0.32       \\ \hline
			3x                     & 66.67     & 3.196    &  0.17       & 2.8       &  0.16      \\ \hline 
			\multicolumn{6}{|c|}{\textbf{VGG-19}} \\ \hline
			1x (VGG XNOR-Net)         & 37.47         & 24.13  &  1         & 24.08    &    1      \\ \hline
			2x        & 44.11         & 10.14  &  0.42         & 10.97    &  0.45        \\ \hline
		\end{tabular}%
		
	}
\vspace{-3mm}	
\end{table}
\vspace{-2mm}
\subsection{Hybrid ResNets}
\begin{figure}[t]
	\centering
	\includegraphics[width=3.6in, keepaspectratio]{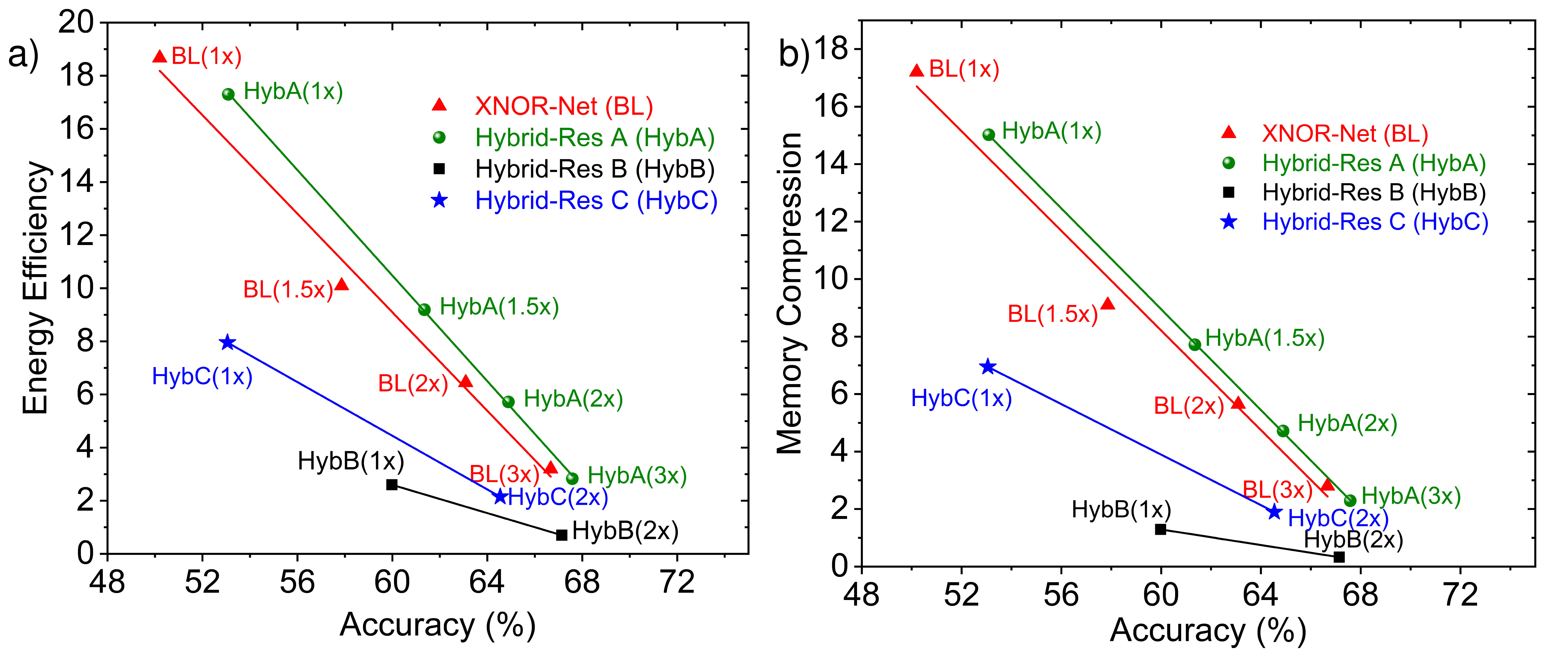}
	\vspace{-6mm}
	\caption{Comparison of different hybrid precision networks and XNOR networks for Resnet-20 architecture in terms of (a) energy-efficiency ((b) Memory compression) and accuracy. The line defining the performance of Hybrid-Res A lies above the lines of inflated networks and Hybrid-Res B and C, thus indicating higher efficiency and compression for iso-performance.}
	\label{fig:hybrid}
	\vspace{-4mm}
\end{figure}

We compare the proposed hybrid ResNet architectures, namely Hybrid-Res A, Hybrid-Res B and Hybrid-Res C. Hybrid-Res A is a XNOR ResNet with full-precision residual connections. Hybrid-Res B consists of 6 full-precision layers and 12 binary layers. Hybrid-Res C has a full-precision fraction of $p=0.1$ in each layer. These numbers are chosen to maintain reasonable compression with respect to full-precision networks. We explore these network architectures for varying network width. Table \ref{Hybrid} lists the accuracy and other metrics for both the hybrid networks and Fig. \ref{fig:hybrid} (a) (and (b)) shows the comparison plot of both XNOR and explored hybrid ResNet architectures in terms of energy efficiency (and memory compression) and accuracy.

\begin{table}[h!]
	\caption{Performance of hybrid ResNet-20 on CIFAR-100}
	\label{Hybrid}
	\vspace{-0.3cm}
	\text{Full-precision Accuracy - 65.81\%}
	\resizebox{0.46\textwidth}{!}{%
		\begin{tabular}{|p{1.5cm}|p{1.5cm}|p{1.5cm}|p{1.5cm}|p{1.5cm}|p{1.5cm}|}
			\hline
			\multicolumn{6}{|c|}{\textbf{Hybrid-Res A (FP Residual)}}                                                  \\ \hline
			Network Width & Accuracy (\%) & E.E (FP) & E.E (XNOR) & M.C (FP) & M.C (XNOR)  \\ \hline
			1x            & 53.09          & 17.3   &   0.93        & 15.023    &   0.87      \\ \hline
			1.5x          & 61.35         & 9.19    &   0.49       & 7.72       &   0.44     \\ \hline
			2x            & 64.89         & 5.72    &   0.3       & 4.72       &   0.27     \\ \hline
			\multicolumn{6}{|c|}{\textbf{Hybrid-Res B (Sectioned Inter-layer)}}                                                  \\ \hline
			Network Width & Accuracy (\%) & E.E (FP) & E.E (XNOR) & M.C (FP) & M.C (XNOR) \\ \hline
			1x            & 59.98           & 2.6    &  0.14         & 1.3       &   0.08      \\ \hline
			2x            & 67.14          & 0.7     &  0.04        & 0.33       &   0.02    \\ \hline
			\multicolumn{6}{|c|}{\textbf{Hybrid-Res C (Sectioned Intra-layer)}}                                                  \\ \hline
			Network Width & Accuracy (\%) & E.E (FP) & E.E (XNOR) & M.C (FP) & M.C (XNOR) \\ \hline
			1x            & 53.05           & 7.96    &  0.43         & 6.95       &   0.4      \\ \hline
			2x            & 64.54          & 2.15     &  0.11        & 1.9       &   0.11    \\ \hline
			
		\end{tabular}%
	}
\vspace{-5mm}
\end{table}

We observe that the hybrid network with full-precision residual connections, Hybrid-Res A, achieves superior tradeoff in terms of accuracy, energy-efficiency and memory compression when hybridizing ResNets compared to sectioned hybrid networks such as Hybrid-Res B and Hybrid-Res C. In fact, Fig. \ref{fig:hybrid} shows that Hybrid-Res A is even superior to just inflating the network width. The results highlights the importance of full-precision residual connections during binarization. 

\vspace{-3mm}

\subsection{Hybrid VGG networks}
Full-precision residual connections offer the optimal way of hybridizing ResNets. We apply the same concept to hybridize VGG networks where we propose a hybrid VGG network by adding full-precision residual connections to a binary VGG network, namely, Hybrid-VGG A. We also explored sectioned hybrid networks described in Section 2.4, Hybrid-VGG B and Hybrid-VGG C to identify the optimum hybrid VGG architecture. Note, Hybrid-VGG B has a full-precision fraction of $p=0.1$ in each layer. Hybrid-VGG C has the linear layers full-precision and the rest binary. These numbers are chosen to maintain reasonable compression with respect to full-precision networks. Table. \ref{VGG} compares the networks explored.

\begin{figure}[t]
	\centering
	\includegraphics[width=3.6in, keepaspectratio]{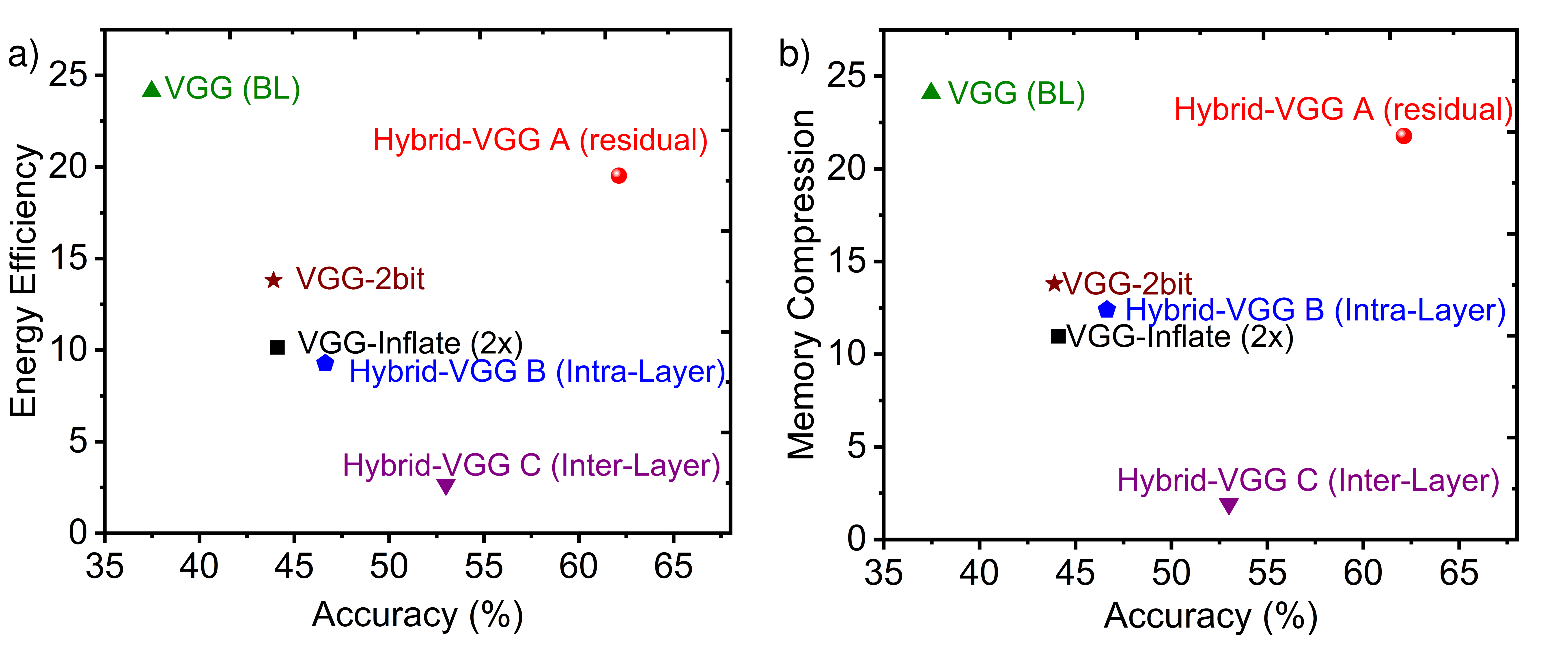}
	\vspace{-0.6cm}
	\caption{Comparison of different hybrid networks for VGG-19 architecture in terms of (a) energy-efficiency and (b) Memory compression v/s accuracy. The proposed Hybrid-VGG A achieves the higher accuracy than other hybrid architectures while achieving 19.53 $\times$ the energy efficiency and 21.78 $\times$ memory compression compared to a full-precision VGG. }
	\vspace{-0.4cm}
	\label{fig:vgg}
\end{figure}

Fig. \ref{fig:vgg} (a) ( and (b)) shows the comparison of networks in terms of accuracy and energy-efficiency (and memory-compression). We observe that the hybrid VGG network with full-precision residual connections (Hybrid-VGG A) achieves the best accuracy while achieving the highest compression and energy efficiency compared to other hybrid networks. We show that Hybrid-VGG A can improves the performance of a VGG XNOR baseline by 25 \% while achieving $19.5\times$ the energy efficiency and $21.8 \times$ the memory compression compared to a full-precision network. Sectioning the network such that the final 3 fully connected layers are full-precision, as in Hybrid-VGG C, do not match Hybrid-VGG A in terms of accuracy despite losing significantly in efficiency and compression. Interestingly, Hybrid-VGG A performs better than Hybrid-VGG B, despite the latter being an alternative network where full-precision information is carried in parallel. To summarize, Hybrid-VGG A emerge as the most optimum hybrid network which is also significantly superior to VGG-2bit and inflated binarized VGG network. These results further show that hybrid networks with full-precision residual connections offer the best tradeoff for improving the performance extremely quantized networks.
\begin{table}[h!]
	\caption{Comparison of hybrid VGG networks on CIFAR-100}
	\label{VGG}
	\vspace{-0.3cm}
	\text{Full-precision Accuracy - 67.21\%}
	\resizebox{0.5\textwidth}{!}{%
		\begin{tabular}{|p{2.1cm}|p{1.5cm}|p{1.5cm}|p{1.5cm}|p{1.5cm}|p{1.6cm}|}
			\hline
			Network Configuration & Accuracy (\%) & E.E (FP) & E.E (XNOR) & M.C (FP) & M.C (XNOR) \\ \hline
			VGG XNOR (BL)         & 37.47         & 24.13  &  1         & 24.08    &    1      \\ \hline
			\textbf{Hybrid-VGG A (residual)}  & \textbf{62.11}         & \textbf{19.53}  &  \textbf{0.81}         & \textbf{21.78}    &   \textbf{0.9}       \\ \hline
			Hybrid-VGG B (Intra-layer) & 46.63         & 9.28   &  0.39         & 12.4     &   0.51       \\ \hline
			Hybrid-VGG C (Inter-layer) & 52.99         & 2.66   &  0.11         & 1.92     &  0.08        \\ \hline
			VGG-2bit     & 43.9          & 13.81  &  0.57         & 13.8     &  0.57        \\ \hline
			VGG-Inflate (2x)        & 44.11         & 10.14  &  0.42         & 10.97    &  0.45        \\ \hline
			
		\end{tabular}
	}
	\vspace{-3mm}
\end{table}
\vspace{-0.3cm}
\begin{table}[h!]
	\caption{Comparison of hybrid VGG networks on ImageNet}
	\label{Imnet}
	\vspace{-0.35cm}
	\text{Full-precision Accuracy 67.69\%}
	\resizebox{0.5\textwidth}{!}{%
		\begin{tabular}{|p{2.2cm}|p{1.5cm}|p{1cm}|p{1.5cm}|p{1.1cm}|p{1.6cm}|}
			\hline
			Network Configuration                       & Accuracy (\%) & E.E (FP) & E.E (XNOR) & M.C (FP) & M.C (XNOR) \\ \hline
			VGG XNOR (BL)                               & 37.72         & 25.56   &    1      & 29.75    &    1      \\ \hline
			Hybrid-VGG A (residual)                        & 48.58         & 17.01   &   0.66       & 28.27    &   0.95       \\ \hline
			Hybrid-VGG C (Linear fp)                       & 57.89         & 6.085   &   0.24       & 1.15     &   0.04       \\ \hline
			\textbf{Hybrid-VGG A' (residual)} & \textbf{57.53}          & \textbf{13.1}    &   \textbf{0.51}       & \textbf{6.6}      &   \textbf{0.22}       \\ \hline
		\end{tabular}
	}
\end{table}
\vspace{-4mm}
\subsection{Scaling to ImageNet}
The dataset ImageNet \cite{deng2009imagenet} is the most challenging dataset pertaining to image classification tasks which is used by the competition ImageNet Large-Scale Visual Recognition Challenge (ILSVRC) to perform the experiments. This subset consists of 1.2 million training images and 50000 validation images divided into 1000 categories. We consider the network VGG-19 for our evaluation and compare Hybrid-VGG A with the VGG-19 XNOR baseline and Hybrid-VGG C. For further comparison, we use a variant of Hybrid-VGG A, described in Section 2.3, named Hybrid-VGG A'. Table. \ref{Imnet} lists the accuracy, energy efficiency and memory compression for the explored networks. We observe that although Hybrid-VGG A does not enhance the performance signficantly, using an additional full-precision linear layer as in Hybrid-VGG A' can increase the performance of the XNOR-Net baseline by $\sim$ 20 \% while still achieving  13.1 $\times$ the energy efficiency compared to full-precision networks. Hybrid-VGG C matches the performance of Hybrid-VGG A', however, at the cost of compression. The results show that the superiority of hybrid networks with full-precision residual connections over other hybrid networks hold true even for larger datasets.
\vspace{-0.1cm}
\vspace{-2mm}
\subsection{Discussion}
%We have proposed hybrid network architectures with full-precision residual connections to achieve reasonable accuracy while still achieving significant energy efficiency and memory compression that binary networks inherently offer. For comparison, we explore sectioned hybrid network architectures where we perform inter-layer and intra-layer sectioning of neural networks into full-precision and binary sections. We analyze the trade-offs between performance, energy efficiency and memory compression. We show that the proposed hybrid architectures with full-precision residual connections offer superior performance compared to the other kinds of hybrid networks we have explored. %
Hybrid networks with full-precision residual connections achieve superior performance in comparison to other hybrid network architectures that were explored in this work. Residual connections offer a parallel path of carrying information from input to output. The hybrid networks with full-precision residual connections exploit this characteristic to partially preserve information lost due to quantization. Residual connections are also computationally simple and the number of weight layers in the residual path is small. Due to this low overhead, using full-precision residual connections in binary networks is a promising technique to match the  performance of full-precision networks while still achieving significant compression and energy efficiency.

The humongous computing power and memory requirements of deep networks stand in the way of ubiquitous use of AI for performing on-chip analytics in low-power edge devices. Memory compression along with the close match to state-of-art accuracies offered by the hybrid extremely quantized networks go a long way to address that challenge. The significant energy efficiency offered by the compressed hybrid networks increases the viability of using AI, powered by deep neural networks, in edge devices. With the proliferation of connected devices in the IOT environment, AI-enabled edge computing can reduce the communication overhead of cloud computing and augment the functionalities of the devices beyond primitive tasks such as sensing, transmission and reception to in-situ processing.

\vspace{-3mm}
\section{Conclusion}
Binary neural networks suffer from  significant degradation in accuracy for deep networks and larger datasets. In this work, we propose extremely quantized hybrid networks with both binary and full-precision sections to closely match full-precision networks in terms of classification accuracy while still achieving significant energy efficiency and memory compression. We explore several hybrid network architecture such as binary networks with full-precision residual connections and sectioned hybrid networks to explore the tradeoffs between performance, energy efficiency and memory compression. Our analysis on ResNets and VGG networks on datasets such as CIFAR-100 and ImageNet show that the hybrid networks with full-precision residual connections emerge as the optimum in terms of accuracy, energy efficiency and memory compression compared to other hybrid networks. This work sheds light on effective ways of designing compressed neural network architectures and potentially paves the way toward using energy-efficient hybrid networks for AI-based on-chip analytics in low-power edge devices with accuracy comparable to full-precision networks.
\section*{Acknowledgement}
This work was supported in part by the Center for Brain-inspired Computing Enabling Autonomous Intelligence (C-BRIC), one of six centers in JUMP, a Semiconductor Research Corporation (SRC) program sponsored by DARPA, in part by the National Science Foundation, in part by Intel, in part by the ONR-MURI program and in part by the Vannevar Bush Faculty Fellowship.
% end the environment with {table*}, NOTE not {table}!

%\printbibliography

\bibliography{acmart}

\begin{thebibliography}{19}
\providecommand{\natexlab}[1]{#1}
\providecommand{\url}[1]{\texttt{#1}}
\expandafter\ifx\csname urlstyle\endcsname\relax
  \providecommand{\doi}[1]{doi: #1}\else
  \providecommand{\doi}{doi: \begingroup \urlstyle{rm}\Url}\fi

\bibitem[Krizhevsky et~al.(2012)]{krizhevsky2012imagenet}
Krizhevsky et~al.
\newblock Imagenet classification with deep convolutional neural networks.
\newblock In \emph{Advances in neural information processing systems}, pages
  1097--1105, 2012.

\bibitem[Szegedy et~al.(2015)]{szegedy2015going}
Szegedy et~al.
\newblock Going deeper with convolutions.
\newblock In \emph{Proceedings of the IEEE conference on computer vision and
  pattern recognition}, pages 1--9, 2015.

\bibitem[He et~al.(2016)]{he2016deep}
He et~al.
\newblock Deep residual learning for image recognition.
\newblock In \emph{Proceedings of the IEEE conference on computer vision and
  pattern recognition}, pages 770--778, 2016.

\bibitem[Girshick et~al.(2015)]{girshick2015fast}
Girshick et~al.
\newblock Fast r-cnn.
\newblock In \emph{Proceedings of the IEEE international conference on computer
  vision}, pages 1440--1448, 2015.

\bibitem[Mikolov et~al.(2013)]{mikolov2013efficient}
Mikolov et~al.
\newblock Efficient estimation of word representations in vector space.
\newblock In \emph{Proceedings of Workshop at International Conference on
  Learning Representations}, 2013.

\bibitem[Mnih et~al.(2015)]{mnih2015human}
Mnih et~al.
\newblock Human-level control through deep reinforcement learning.
\newblock \emph{Nature}, 518\penalty0 (7540):\penalty0 529, 2015.

\bibitem[Silver et~al.(2017)]{silver2017mastering}
Silver et~al.
\newblock Mastering the game of go without human knowledge.
\newblock \emph{Nature}, 550\penalty0 (7676):\penalty0 354, 2017.

\bibitem[Iandola et~al.(2016)]{iandola2016squeezenet}
Iandola et~al.
\newblock Squeezenet: Alexnet-level accuracy with 50x fewer parameters and< 0.5
  mb model size.
\newblock \emph{arXiv preprint arXiv:1602.07360}, 2016.

\bibitem[Alvarez and Salzmann(2017)]{alvarez2017compression}
Alvarez and Salzmann.
\newblock Compression-aware training of deep networks.
\newblock In \emph{Advances in Neural Information Processing Systems}, pages
  856--867, 2017.

\bibitem[Weigend et~al.(1991)]{weigend1991generalization}
Weigend et~al.
\newblock Generalization by weight-elimination with application to forecasting.
\newblock In \emph{Advances in neural information processing systems}, pages
  875--882, 1991.

\bibitem[Han et~al.(2015)]{han2015learning}
Han et~al.
\newblock Learning both weights and connections for efficient neural network.
\newblock In \emph{Advances in neural information processing systems}, pages
  1135--1143, 2015.

\bibitem[Ullrich et~al.(2017)]{ullrich2017soft}
Ullrich et~al.
\newblock Soft weight-sharing for neural network compression.
\newblock \emph{arXiv preprint arXiv:1702.04008}, 2017.

\bibitem[Hubara et~al.(2017)]{hubara2017quantized}
Hubara et~al.
\newblock Quantized neural networks: Training neural networks with low
  precision weights and activations.
\newblock \emph{The Journal of Machine Learning Research}, 18\penalty0
  (1):\penalty0 6869--6898, 2017.

\bibitem[Mellempudi et~al.(2017)]{mellempudi2017ternary}
Mellempudi et~al.
\newblock Ternary neural networks with fine-grained quantization.
\newblock \emph{arXiv preprint arXiv:1705.01462}, 2017.

\bibitem[Rastegari et~al.(2016)]{rastegari2016xnor}
Rastegari et~al.
\newblock Xnor-net: Imagenet classification using binary convolutional neural
  networks.
\newblock In \emph{European Conference on Computer Vision}, pages 525--542.
  Springer, 2016.

\bibitem[Prabhu et~al.(2018)]{prabhu2018hybrid}
Prabhu et~al.
\newblock Hybrid binary networks: Optimizing for accuracy, efficiency and
  memory.
\newblock In \emph{2018 IEEE Winter Conference on Applications of Computer
  Vision (WACV)}, pages 821--829. IEEE, 2018.

\bibitem[Keckler et~al.(2011)]{keckler2011gpus}
Keckler et~al.
\newblock Gpus and the future of parallel computing.
\newblock \emph{IEEE Micro}, \penalty0 (5):\penalty0 7--17, 2011.

\bibitem[Krizhevsky and Hinton(2009)]{krizhevsky2009learning}
Krizhevsky and Hinton.
\newblock Learning multiple layers of features from tiny images.
\newblock Technical report, Citeseer, 2009.

\bibitem[Deng et~al.(2009)]{deng2009imagenet}
Deng et~al.
\newblock Imagenet: A large-scale hierarchical image database.
\newblock In \emph{Computer Vision and Pattern Recognition, 2009. CVPR 2009.
  IEEE Conference on}, pages 248--255. Ieee, 2009.

\end{thebibliography}
\bibliographystyle{IEEEtran}

\end{document}